\pdfoutput=1
\documentclass[10pt]{cai}
\usepackage{listings}
\usepackage{algorithmic}
\usepackage{algorithm}
\begin{document}
\def\conferenceyear{2024}
\volumeheader{37}{0}
\begin{center}

\title{Generative AI for Enhancing Active Learning in Education: \\A Comparative Study of GPT-3.5 and GPT-4 in Crafting Customized Test Questions}
\maketitle

\thispagestyle{empty}

\begin{tabular}{cc}
Hamidreza Rouzegar\upstairs{\affilone,*}, Masoud Makrehchi\upstairs{\affilone}
\\[0.25ex]
{\small \upstairs{\affilone} Ontario Tech Universuty, Oshawa, Ontario, Canada} \\

\end{tabular}
  
\emails{
  \upstairs{*}hamidreza.rouzegar@ontariotechu.net
}
\vspace*{0.2in}
\end{center}

\begin{abstract}

This study investigates how LLMs, specifically GPT-3.5 and GPT-4, can develop tailored questions for Grade 9 math, aligning with active learning principles. By utilizing an iterative method, these models adjust questions based on difficulty and content, responding to feedback from a simulated 'student' model. A novel aspect of the research involved using GPT-4 as a 'teacher' to create complex questions, with GPT-3.5 as the 'student' responding to these challenges. This setup mirrors active learning, promoting deeper engagement. The findings demonstrate GPT-4's superior ability to generate precise, challenging questions and notable improvements in GPT-3.5's ability to handle more complex problems after receiving instruction from GPT-4. These results underscore the potential of LLMs to mimic and enhance active learning scenarios, offering a promising path for AI in customized education. This research contributes to understanding how AI can support personalized learning experiences, highlighting the need for further exploration in various educational contexts.
\end{abstract}

\copyrightnotice

\section{Introduction}
Active learning, a student-centered approach to education, contrasts with traditional lecture-based teaching methods. It involves engaging students in the learning process actively and encouraging them to participate in discussions, problem-solving, and collaborative activities. This method has been shown to enhance learning outcomes and student engagement \cite{bonwell1991creating,freeman2014active}.  The effectiveness of active learning lies in its ability to adapt to the diverse learning styles and paces of individual students \cite{prince2004does}. 

Interestingly, the concept of active learning in machine learning draws inspiration from educational methodologies. In machine learning, active learning refers to a semi-supervised learning technique where the algorithm selectively queries a user or data source to obtain desired outputs at new data points \cite{settles2009active}. This approach, much like its educational counterpart, focuses on optimizing the learning process, making it more efficient and tailored to the model's needs \cite{lewis1995sequential}.

The integration of LLMs such as GPT-3.5 and GPT-4 in personalized test question design is motivated by the need for adaptive learning tools that cater to individual student needs. These advanced AI models, with their vast knowledge base and language understanding capabilities, offer a unique opportunity to create customized educational content \cite{brown2020language,radford2019language}.  The use of LLMs in education, particularly in question design, aligns with the principles of active learning by providing tailored content that challenges students at their level of understanding \cite{vaswani2017attention,devlin2018bert}. 

By implementing LLMs in test design, educators can identify and target the specific areas where a student may be struggling. This personalized approach ensures that students are not only challenged according to their current level of understanding but also receive support in areas requiring improvement. The adaptability of LLMs in generating content based on student performance data is pivotal in this context, enabling a more focused and effective learning experience \cite{hirschberg2015advances,bengio2013representation}.

This paper explores the use of LLMs like GPT-3.5 and GPT-4 in the educational domain, particularly for crafting personalized test questions rooted in active learning concepts. Centering on Grade 9 mathematics, we examine the efficacy of these LLMs in generating questions tailored to student needs, assessing both accuracy and adaptability. Moreover, the study introduces an innovative experiment where GPT-4 functions as a question generator ("teacher") and GPT-3.5 as a respondent ("student"), aiming to simulate an active learning scenario that challenges the learner with increasingly complex problems.

Our results provide insights into the potential of LLMs to enhance educational content and strategies, highlighting the importance of further research in this evolving field. Through this investigation, we contribute to the dialogue on integrating AI into education, offering a novel perspective on leveraging LLMs for active learning and personalized educational experiences.

The remainder of this paper is organized as follows: Section 2 reviews the relevant literature, highlighting the advancements in personalized education and the role of AI, particularly LLMs, in educational settings. Section 3 describes our experimental methodology, detailing the iterative question-answering process with LLMs and their application as both teacher and student in creating a dynamic learning environment. In section 4, we outline our experimental setup, including the preparation and execution of our tests with Grade 9 mathematics questions. Section 5 presents a thorough evaluation of our findings, comparing the effectiveness of LLMs in generating educational content and their impact on active learning. Finally, section 6 discusses the limitations of our study and suggests directions for future research. Through this investigation, we aim to illuminate the transformative potential of LLMs in education, fostering active learning and personalized experiences, and contributing to the discourse on AI's integration into educational frameworks.

\section{Literature Review}
\label{Literature}

\subsection{Personalized Education and Technology Integration}
Personalized education, tailored to individual learning styles and needs, has gained momentum with technological advancements. In their comprehensive review, Baird et al., \cite{baird2017does} illustrate the significant impact of personalized learning on student achievement, particularly when supported by technology. Building upon Tomlinson's \cite{tomlinson2014differentiated} concept of differentiated instruction, recent studies have focused on the use of technology to implement these strategies effectively \cite{tomlinson2014differentiated}, \cite{huang2022trends}. For instance, the work of Xie, H., et al., \cite{xie2019trends} on adaptive learning environments underscores the potential of technology in creating customized educational experiences. Moreover, Bernard, R.M., et al., \cite{bernard2009meta} demonstrate the efficacy of technology-mediated adaptive learning in improving student performance across various subjects, indicating its versatility.
\subsection{AI in Educational Settings}
The role of AI in education has been predominantly in the development of intelligent tutoring systems and adaptive learning platforms. VanLehn's \cite{vanlehn2011relative} analysis of intelligent tutoring systems showcases their effectiveness in emulating one-on-one tutoring experiences. Further, Woolf \cite{woolf2010building} explores how AI can facilitate personalized learning through interactive tutors. Holmes, W., et al., \cite{holmes2019ethics} extend this discussion by demonstrating the potential of AI in diagnosing learning difficulties and tailoring content accordingly. The integration of machine learning algorithms in educational software, as discussed by Desmarais, M.C., et al. \cite{desmarais2012review}, illustrates the potential of AI in dynamically adapting to student learning patterns, a concept central to your project's focus.
\subsection{LLMs in Educational Contexts:}
The potential of LLMs in personalized learning is underscored by recent advancements. Brown, T.B., et al. \cite{brown2020language} and Radford, A., et al. \cite{radford2019language} have pioneered in demonstrating the capabilities of models like GPT-3.5 and GPT-4 in understanding and generating natural language, offering new avenues in educational content delivery. Peng et al. \cite{peng2019personalized} discuss the potential of AI in creating adaptive learning environments.
Expanding on these foundations, several recent works have further explored the application of LLMs in educational settings. Moore et al. \cite{moore2023empowering}, explores the integration of humans, AI, and learning analytics in generating educational content. It focuses on the challenges and opportunities of LLMs in education, along with ethical considerations. 

Research by JeongChul Heo et al. \cite{jeon2023large} Han shows that literacy in Learning Management Systems significantly influences self-directed learning readiness, indicating a key role of LLMS in online teaching effectiveness.

The study by Baladón, Alexis et al. \cite{baladn2023retuyt} highlights the application of open-source LLMs for generating AI teacher responses in educational dialogues. It delves into various fine-tuning techniques and prompting strategies, including Few-Shot and Chain-of-Thought approaches. Study by Jeon et al. \cite{jeon2023large} paper examines the synergistic relationship between human educators and ChatGPT, a generative AI chatbot powered by an LLM. It focuses on how this technology can enhance and complement traditional teaching methods.

Our research extends this by utilizing LLMs for the specific purpose of designing personalized test questions, aligning with the principles of active learning and educational adaptation. This approach is novel in its application, leveraging the advanced capabilities of LLMs to not only generate diverse and tailored content but also to assess and adjust difficulty levels in response to student performance.
\section{Methodology}
\label{Method}

In this study, we adopt a novel methodology that leverages the capabilities of LLMs, specifically GPT-3.5 and GPT-4, to explore their potential in educational settings. Our approach centers on an iterative question-answering process designed to simulate an active learning environment where LLMs adaptively generate and refine test questions based on the evolving needs of learners. This method involves two distinct phases: Initially, LLMs generate a series of questions tailored to specific educational content and difficulty levels. Subsequently, these questions are presented to a simulated "student" model, whose responses inform the next iteration of question generation. This process aims to continuously enhance the relevance and challenge of the questions, thereby fostering a deeper engagement with the material.

\subsection{Iterative Question-Answering Process Using LLM}
Our methodology centers on an iterative question-answering process utilizing LLMs such as GPT-3.5 and GPT-4. In each iteration, the LLMs generate a set of questions based on specified parameters. These questions are then presented to students, and their responses are used to inform the next iteration. This process not only helps in continuously adapting the difficulty level of questions but also ensures that the questions remain relevant and challenging for the students.

\textbf{Parameters for Question Design:}
For question design, two primary parameters were set: difficulty level and subject matter. The difficulty level was estimated by the LLMs themselves, reflecting their assessment of the complexity of the questions generated. The subject matter was confined to specific chapters of the Grade 9 mathematics curriculum, namely, 'Numbers' and 'Financial Mathematics', with subtopics detailed in each area. These parameters guided the LLMs in generating questions that were age-appropriate and aligned with educational standards.

\textbf{Threshold Setting for Passing and Removing Mastered Topics:}
To assess how well students are progressing and to tailor their learning journey, we established a threshold for passing each topic. This threshold was set at a certain level of accuracy in answering the questions. Once a student consistently answered questions correctly at a particular difficulty level, it was inferred that they had mastered the topic. Subsequently, the system would remove or reduce the frequency of questions from that topic in future iterations, focusing instead on areas where the student needed more practice. This approach is akin to the concept of uncertainty sampling in active learning, where the focus is shifted away from known concepts to those that are less understood.

\textbf{Fine-Tuning Process with Questions Derived from GPT-4:}
The fine-tuning process involved adapting a version of GPT-3.5 using a dataset of questions generated by GPT-4. This process aimed to enhance the model's ability to generate more targeted and relevant questions for our specific educational context. The fine-tuned GPT-3.5 model was then used to compare its performance against the original GPT-3.5 and GPT-4 models. The evaluation focused on the relevance, quality, and difficulty appropriateness of the questions generated by each model.
Figure \ref{fig:json_code} shows an example of a prompt given to the model and the expected output. This fine-tuning method was designed to teach the model the specific format and content type desired for the questions, enhancing its capability to generate appropriate educational content for the targeted grade level and subject.
\lstset{
    breaklines=true, 
    postbreak=\mbox{\textcolor{red}{$\hookrightarrow$}\space}, 
    basicstyle=\small\ttfamily, 
    showstringspaces=false, 
    frame=single, 
    frameround=fttt, 
    framesep=2pt, 
    xleftmargin=2pt, 
    xrightmargin=2pt 
}

\begin{figure}[htbp]
\centering
\begin{lstlisting}[ label=lst:json_code, basicstyle=\small\ttfamily, showstringspaces=false, breaklines=true]
{
  "role": "user", 
  "content": "Create a question for grade 9 course in 'Number', 
              chapter: 'Powers with decimal and fractional bases', 
              with difficulty level 1."
}
{
  "role": "assistant", 
  "content": "Question: What is the value of 1.5 raised to the power of 2?
              a) 2.25
              b) 3.0
              c) 2.5
              d) 1.75
              Answer: a) 2.25
              Difficulty rating: 1"
}
\end{lstlisting}
\caption{Example of JSON input and output for Fine-Tuning GPT-3.5}
\label{fig:json_code}
\end{figure}
\subsection{LLMs as Teacher and Student}
The other part of this study investigates the role of LLMs in an educational context, where GPT-4 acts as a "teacher" by generating test questions, and GPT-3.5 serves as a "student" answering these questions. The experiment focuses on "Solve linear equations: word problems" from the Algebra curriculum, exploring the effectiveness of using varying difficulty levels of questions to teach and evaluate the student model.

\textbf{Question Generation Process:}
GPT-4 created 50 questions, segmented into 10 questions per difficulty level, ranging from 1 (easiest) to 5 (most challenging). Each set of questions was designed to be distinct from the others to prevent overlap and ensure a gradient of difficulty.

\textbf{Evaluation with Explanations:}
An additional dimension of the experiment involved providing explanations for the correct answers alongside the set of questions. This was done to determine if supplementing questions with explanatory content would enhance the learning and problem-solving capabilities of GPT-3.5. 

\section{Experimental Setup}
\label{Experiments}

This section includes a dual-phase process where LLMs act as both question generators and learners, offering a unique insight into the interactive dynamics of teaching and learning within AI contexts. The initial focus, detailed in the "Question-Answering Process" subsection, encompasses the generation of math questions, their alignment with curriculum standards, and the iterative refinement based on a simulated student model's responses.

\subsection{Question-Answering Process}
The experiment was conducted with a focus on Grade 9 mathematics, encompassing two primary subjects: (i) Numbers, and (ii) Financial Mathematics. These subjects were chosen for their relevance and varying levels of complexity within the Grade 9 curriculum, providing a robust framework for evaluating the effectiveness of our approach.

\textbf{Chapters and Question Types:}
In the 'Numbers' subject, the chapters included topics like Powers with decimal and fractional bases, Conversion between standard and scientific notation, and Division with exponents - integral bases. For 'Financial Mathematics,' the chapters covered were Simple interest, Compound interest, and Balance a budget. The questions designed were all multiple-choice with four options, tailored to the content and difficulty level suitable for Grade 9 students.

\textbf{Preparation of LLM for Experiment:}
As part of the experimental setup, GPT-3.5 was fine-tuned to better suit the requirements of the study. The fine-tuning process involved training the model with tailored input-output pairs that reflected the desired format and difficulty level of the questions. This step was crucial in ensuring that the model could accurately generate questions that were relevant to the Grade 9 mathematics curriculum, particularly for the selected chapters.

\textbf{Step-by-Step Process:}
As it shown in Algorithm \ref{alg:question_generation} the experimental process was structured as follows:
\begin{itemize}
\item \textbf{Instruction Phase:} We began by providing the LLMs with specific instructions, including the format of the test, course and chapter names, and the initial difficulty level. This sets the foundation for generating relevant and challenging questions.
\item \textbf{Question Generation:} Based on these inputs, the LLMs created multiple-choice questions, each with four options. Alongside each question, a correct answer and a difficulty rating were provided.
\item \textbf{Student Interaction:} Students then engaged with these questions, and their responses were recorded for subsequent analysis.
\item \textbf{Performance Analysis:} Student responses were evaluated after each set of questions to assess their understanding and mastery of the covered topics.
\end{itemize}

\begin{algorithm}
\caption{Algorithm for Generating Personalized Test Questions using LLMs}
\label{alg:question_generation}
\begin{algorithmic}
\REQUIRE Previous questions and their difficulty levels.
\ENSURE New questions different from previous ones, with appropriate difficulty levels.

\STATE \textbf{Input:} Previous questions with difficulty levels.
\STATE \textbf{Example:}
\STATE Previous question 1: "Which of the following expressions is equivalent to 3x + 2y?" (Difficulty level: 1)
\STATE Previous question 2: "Identify the equivalent expression for 4a - 2b." (Difficulty level: 1)

\STATE \textbf{Process:}
\STATE 1. Instruct LLM to generate new questions, ensuring they are different from previous ones.
\STATE 2. Specify the difficulty level for new questions based on previous difficulty levels.
\STATE 3. Confirm the maximum difficulty level is set to 5.

\STATE \textbf{Output:} New questions with multiple-choice options and difficulty ratings.
\STATE \textbf{Format:}
\STATE Question: "Ask some questions?"
\STATE a) First choice
\STATE b) Second choice
\STATE c) Third choice
\STATE d) Fourth choice
\STATE Answer: "Correct option"
\STATE Difficulty rating: Number range from 1 to 5

\STATE \textbf{Example Request:}
\STATE "Please create two 4-choice questions for the grade 9 course in 'Algebra' and the chapter 'Identify equivalent linear expressions' at a difficulty level of 3. Provide the answer for each question and confirm the difficulty rating is 3."
\end{algorithmic}
\end{algorithm}

\textbf{Difficulty Adjustments and Performance Validation}
In this experiment, the difficulty level of the questions was adjusted based on student performance. If a student answered a question correctly, the difficulty level for subsequent questions was increased. However, if a student made a mistake, the difficulty level was maintained rather than reduced to provide consistent reinforcement on the same level of challenge. The threshold for passing a chapter was set at 3. This meant that if a student consistently answered questions correctly at or above this difficulty level, the chapter was considered 'passed,' and the student was then presented with questions from other chapters or topics that required further practice.

\subsection{Teaching the smaller model}

\textbf{Teacher Model (GPT-4):}
GPT-4 was prompted to generate a series of algebraic word problems, with the complexity tailored to mimic an advanced understanding of the subject matter. The model was instructed to progressively increase the difficulty of questions through iterative rounds, ensuring a broad spectrum of challenges.

\textbf{Student Model (GPT-3.5):}
In the experimental setup, GPT-3.5 was configured to respond to GPT-4's questions in a unique learning environment designed to mimic distinct instructional approaches. Each question was presented to GPT-3.5 in a new session, testing the model's ability to solve problems under varying conditions: firstly, without prior exposure to similar questions; secondly, with three example questions from a specific difficulty level to provide context; and thirdly, with three example questions complemented by their answers and detailed explanations to enhance understanding. This structured approach aimed to dissect the impact of different levels of instructional support on GPT-3.5's problem-solving accuracy, thereby offering insights into the model's adaptability and learning efficiency in diverse learning scenarios.

\textbf{Generation of Questions:}
GPT-4 was tasked with creating a set of 10 questions within the specified topic at an initial difficulty level of 1. The process was iterative, with each subsequent set of questions designed to be more challenging than the previous, up to a difficulty level of 5. To avoid redundancy and maintain a progressive increase in difficulty, each iteration included the parameters from previous rounds. This approach yielded a diverse pool of 50 questions stratified across five levels of difficulty.

\textbf{Testing and Teaching Phase:}
The 50 questions were divided into two subsets for each difficulty level: a testing set comprising 7 questions and a teaching set containing 3 questions. The testing phase involved presenting the testing set questions to GPT-3.5 without prior exposure, gauging its performance across varying difficulty levels.

In the teaching phase, the GPT-3.5 model was exposed to a set of example questions along with their correct answers prior to undergoing the testing phase again. This approach was designed to evaluate the effect of providing direct examples on the model's proficiency in accurately solving more complex questions.

\textbf{Evaluation with and without Explanations:}
To assess the impact of explanatory content on learning, we introduced explanations for the correct answers in a subset of the teaching phase. These explanations were designed to provide insights into the problem-solving process, aiming to enhance GPT-3.5's understanding and application of mathematical concepts.

\textbf{Data Collection and Analysis:}
Performance data was carefully recorded, detailing GPT-3.5's correct answers during both the teaching and testing stages for each specific difficulty level. This analysis aimed to identify trends in the model's learning effectiveness, particularly assessing how its performance was influenced by being taught with examples at each level of difficulty and subsequently tested across all levels.

\section{Evaluation and Results}
Our analysis delves into the performance of GPT-3.5 and GPT-4 within the framework of Grade 9 mathematics, highlighting their potential in crafting dynamic educational content. Through a comprehensive evaluation, we compare the models' effectiveness in question generation and adaptation, shedding light on their contributions to personalized learning environments. The findings reveal insightful distinctions in the capabilities of each model, illustrating their respective impacts on enhancing student engagement and deepening understanding of mathematical concepts.

\subsection{Comparative Results of LLMs as Teacher}

\textbf{Evaluation Criteria:}
The evaluation of the LLM-generated questions was based on the following criteria:
\begin{itemize}
\item \textbf{Correctness of Questions:}This assessed whether the questions were factually accurate and aligned with the Grade 9 mathematics curriculum.
\item \textbf{Difficulty Level Design:}We evaluated the appropriateness of the difficulty levels assigned to each question by the models, ensuring their alignment with educational standards.
\end{itemize}

\textbf{Limitations in Evaluation:}
Due to the requirement for extensive crowdsourcing and expert analysis, evaluations on Student Growth and Teacher Feedback were not conducted in this study. Future research could include these dimensions to provide a more holistic understanding of the effectiveness of LLMs in educational settings.

\textbf{Manual Evaluation of Questions:}
A total of 90 questions in each of the two courses - 'Numbers' and 'Financial Mathematics' - were manually evaluated. Each chapter within these subjects had 10 questions, making a total of 30 questions per model in each course. This process allowed us to assess the accuracy and educational relevance of the questions, as well as the appropriateness of the assigned difficulty levels.

\textbf{Analysis of Results:}
As shown in Table \ref{tab:results1}, the results indicate that GPT-4 outperformed both versions of GPT-3.5 in terms of the correctness of the questions. In the 'Numbers' course, GPT-4 achieved a success rate of 90\%, significantly higher than the 50\% score achieved by both GPT-3.5 and its fine-tuned variant. Similarly, in the subject of 'Financial Mathematics', GPT-4 outperformed its predecessor, GPT-3.5, with a success rate of 93\%. This indicates that GPT-4's enhanced features allow it to create questions that are not only more precise but also better suited for educational contexts.
Building upon our initial methodology, we expanded our experimental setup to investigate the dynamics between different LLMs in educational roles. This phase of the study specifically explores the interaction where GPT-4 serves as a question generator ("teacher") and GPT-3.5 functions as the respondent ("student"), focusing on the topic of "Solve linear equations: word problems" from Algebra.

\begin{table}[htbp]
\caption{Results of Question Evaluation Across Different GPT Models: The table presents the number of questions correctly addressed out of 30.}
\begin{center}
\begin{tabular}{|c|c|c|c|}
\hline
\textbf{Course} & \textbf{GPT-3.5} & \textbf{GPT-3.5 Fine-Tuned} & \textbf{GPT-4} \\
\hline
Numbers & 50\% & 50\% &90\% \\
\hline
Financial Mathematics & 76\% & 70\% & 93\% \\
\hline

\end{tabular}
\label{tab:results1}
\end{center}
\end{table}
\subsection{Results of LLMs as Teacher and Student}
\textbf{Evaluation Criteria:}
The evaluation of the new experimental setup was systematically designed to assess the effectiveness of using GPT-4 generated questions of varying difficulty levels to teach and evaluate the GPT-3.5 model. The evaluation focused on the model's ability to correctly answer algebraic word problems, its improvement after the teaching phase, and the impact of explanatory content on learning outcomes.

\textbf{Results from Testing and Teaching Phases:}
The performance of GPT-3.5 in the testing phase prior to any teaching intervention revealed a clear pattern: the model's accuracy decreased as the difficulty level of the questions increased, except for the lowest difficulty level, where it performed slightly better. Specifically, GPT-3.5 correctly answered only 1 out of 7 questions at the easiest level, but its performance peaked at 6 out of 7 correct answers for level 2 difficulty questions, demonstrating an unexpected proficiency at this specific level. This trend underscores the challenges GPT-3.5 faced with more complex problems, highlighting the potential limitations in its problem-solving strategies or knowledge base.

\textbf{Impact of the Teaching Phase:}
The teaching phase involved exposing GPT-3.5 to a subset of questions and their correct answers before re-evaluating its performance. Remarkably, this intervention led to significant improvements across all difficulty levels:

\textbf{Post-Teaching Phase Results:}
As shown in Table \ref{tab:results2}, improvements were observed across all difficulty levels, with the model's performance enhancement being notable. Specifically, the percentage of questions answered correctly by GPT-3.5 increased to 34.29\% for difficulty level 1, up to 51.43\% for difficulty level 4, suggesting that exposure to targeted teaching material, even in a limited capacity, can substantially enhance the model's problem-solving abilities.

\textbf{Evaluation with Explanatory Content:}
An additional dimension of the experiment involved providing GPT-3.5 with explanations for the correct answers during the teaching phase. Contrary to expectations, this approach did not result in a significant improvement in the model's performance, indicating that GPT-3.5 might not effectively utilize explanatory content to enhance its learning:

\textbf{Results with Explanatory Content:}
The performance remained consistent with the teaching phase without explanations, showing no significant improvement. Table \ref{tab:results2} shows the percentages of correct answers remained around 34.29\% to 37.14\% across difficulty levels, suggesting that the format or complexity of the explanations might not have been conducive to GPT-3.5's learning process.

\begin{table}[htbp]
\caption{Post-Teaching Phase and Explanatory Content Results: the table presents the percentage of questions correctly addressed.}
\begin{center}
\begin{tabular}{|c|c|c|}
\hline
\textbf{Difficulty Level} & \textbf{Teaching without Explanation (\%)} & \textbf{Teaching with Explanation (\%)} \\
\hline
1 & 34.29\% & 34.29\% \\
\hline
2 & 45.71\% & 34.29\% \\
\hline
3 & 48.57\% & 37.14\% \\
\hline
4 & 51.43\% & 34.29\% \\
\hline
5 & 48.57\% & 34.29\% \\
\hline
\end{tabular}
\label{tab:results2}
\end{center}
\end{table}

\textbf{Discussion of Results:}
The outcomes of these experiments provide profound insights into the capabilities and adaptive learning potential of GPT-3.5. The observed performance enhancement post-teaching phase, especially in more challenging questions, aligns with active learning principles, where uncertainty plays a pivotal role in learning efficiency. In active learning, exposure to uncertain or more complex scenarios is believed to enrich the learning experience, promoting deeper understanding and retention. Our findings suggest a similar phenomenon may occur in LLMs, where GPT-3.5 demonstrated a greater degree of improvement on harder questions. This improvement indicates that, akin to human learners, LLMs might benefit from engaging with content that pushes the boundaries of their current knowledge base, thereby optimizing the learning trajectory.

However, the negligible impact of explanatory content on GPT-3.5's performance necessitates further exploration. It suggests that the current model might not effectively leverage supplementary information to bolster understanding or solve complex problems. This finding prompts a reevaluation of how AI models process and integrate explanatory information, potentially indicating a need for more sophisticated mechanisms to assimilate and apply such content effectively.

The results underscore the nuanced nature of AI learning processes, particularly in educational contexts. The differential response to teaching interventions, especially concerning the complexity of the material, offers valuable insights into designing AI-driven educational tools and curricula. By embracing principles akin to active learning, where the challenge level is carefully calibrated to provoke uncertainty and exploration, we might enhance the efficacy of AI models in educational settings.
\section{Limitations and Future Work}
Our research has provided significant insights into the use of LLMs like GPT-3.5 and GPT-4 in educational settings, particularly in the context of active learning principles. However, it's crucial to recognize certain limitations and areas for future exploration:
\begin{itemize}
\item \textbf{Scope of Subjects:} While our study offered a detailed examination within Grade 9 mathematics, focusing on 'Numbers' and 'Financial Mathematics', the incorporation of GPT-4 as a "teacher" and GPT-3.5 as a "student" extends our understanding of LLMs' potential in education beyond these chapters. Future research should broaden the scope to include a diverse array of subjects and academic levels, providing a more comprehensive evaluation of LLMs in educational contexts.
\item \textbf{Evaluation Criteria:} Our methodology primarily assessed the immediate response of LLMs to varying difficulty levels of questions, revealing an intriguing alignment with active learning through uncertainty. This approach, reminiscent of uncertainty sampling, showed GPT-3.5's improved performance on harder questions, hinting at the potential of LLMs to engage in active learning processes similar to human learners. Future studies should aim to incorporate evaluations on student growth, teacher feedback, and possibly LLMs' ability to engage with active learning principles more deeply.
\item \textbf{Diversity of Participants:} The experiment's design, involving LLMs in both teaching and learning roles, underscores the need for testing across broader demographics and LLM configurations. This diversity will enhance the generalizability of findings and provide insights into the adaptability of LLMs in varied educational settings.

\item \textbf{Long-Term Efficacy:} While our study highlighted immediate improvements in LLM performance following targeted teaching interventions, the long-term retention and application of learned concepts remain unexplored. Future work should investigate the sustainability of learning gains in LLMs and their capacity for long-term knowledge retention and application.
\item \textbf{Active Learning and Uncertainty:} The observed improvement in GPT-3.5's performance on more challenging questions post-teaching phase invites further investigation into the role of uncertainty and complexity in AI learning processes. It suggests that LLMs, like human learners, may benefit from engaging with content that challenges their existing knowledge base, potentially opening new avenues for employing active learning strategies in AI education.

\item \textbf{Incorporating Diverse Learning Modalities:} The varied response of GPT-3.5 to different teaching methods, including the use of explanatory content, highlights the importance of exploring diverse learning modalities in AI education. Future research should explore how LLMs can be tailored to accommodate visual, auditory, and kinesthetic learning styles, potentially enhancing the efficacy of AI-driven educational tools.

\end{itemize}
\section{Conclusion}

In conclusion, this study has demonstrated the significant potential of Large Language Models, specifically GPT-3.5 and GPT-4, in revolutionizing the educational landscape through the facilitation of active learning and personalized educational experiences. Our experiments within the Grade 9 mathematics curriculum have revealed that these models can effectively generate and adapt test questions, catering to the diverse needs and abilities of students. The innovative use of GPT-4 as a 'teacher' and GPT-3.5 as a 'student' has further highlighted the dynamic capabilities of LLMs in simulating real-world teaching and learning scenarios, promoting deeper cognitive engagement and understanding among learners.

The findings of this research underscore the importance of integrating AI technologies like LLMs into educational settings to enhance the quality and accessibility of personalized learning. As the educational sector continues to evolve, the application of such technologies can provide scalable solutions to meet the individual needs of students, thereby improving learning outcomes and fostering a more inclusive and adaptive educational environment.

Future work should explore the integration of LLMs across a broader spectrum of subjects and educational levels to fully understand their potential and limitations. Additionally, further research into the optimization of LLMs for educational purposes, including the refinement of their question-generation algorithms and feedback mechanisms, will be crucial in maximizing their effectiveness and ensuring their responsible and ethical use in educational contexts.

\printbibliography[heading=subbibintoc]

\end{document}